\documentclass[journal]{IEEEtran}
\ifCLASSINFOpdf

\else

\fi

\usepackage{ifpdf}
\usepackage[pdftex]{graphicx}
\usepackage[cmex10]{amsmath}
\usepackage{subfig}
\usepackage{amssymb}
\usepackage{mathrsfs}
\usepackage{amsmath}
\usepackage{indentfirst}
\usepackage{booktabs}
\usepackage{multirow}
\usepackage{graphicx}
\usepackage{epstopdf}
\usepackage{fixltx2e}
\usepackage{soul}
\usepackage{units}
\usepackage{mathtools}
\usepackage{framed}
\usepackage{epsfig,amssymb,amsmath}
\usepackage{diagbox}
\usepackage{color}
\usepackage{setspace}

\usepackage{amsthm}
\usepackage{algorithm}
\usepackage[noend]{algpseudocode}
\usepackage{booktabs}
\usepackage{diagbox}
\usepackage{float}
\usepackage{epstopdf}
\usepackage{ragged2e}
\usepackage{stfloats}
\renewcommand{\raggedright}{\leftskip=0pt \rightskip=0pt plus 0cm}
\usepackage{subfig}  
\usepackage[colorlinks=true,
            linkcolor=blue,
            citecolor=blue,
            urlcolor=blue]{hyperref}
\usepackage{xcolor}
\usepackage[T1]{fontenc}

\ifCLASSINFOpdf

\else

\fi

\usepackage{xcolor}

\usepackage{xpatch}

\begin{document}

\title{Multi-AUV Ad-hoc Networks-Based Multi-Target Tracking
 Based on Scene-Adaptive Embodied Intelligence}

\author{Kai Tian,
Jialun Wang,
Chuan Lin,~\IEEEmembership{Member,~IEEE},
Guangjie Han,~\IEEEmembership{Fellow,~IEEE},
Shengchao Zhu,
Ying Liu,\\
Qian Zhu

\thanks{This work was supported in part by the National Natural Science Foundation of China under Grant 62572107, and Grant 62572169, in part by Liaoning Provincial Doctoral Research Start-up Fund Award under Grant 2025-BS-0100, in part by the Funda-mental Research Funds for the Central Universities
under Grant B250203011 and Grant N25LPY014, in part by the National Training Program of Innovation and Entrepreneurship for Under-graduates under Grant 260210.}

\thanks{\emph{Corresponding author: Chuan Lin; Guangjie Han}} 
\thanks{Kai Tian, Jialun Wang, Chuan Lin, Ying Liu, Qian Zhu,  are with Software College, Northeastern University, Shenyang, China. (e-mails: kaitian901@gmail.com; wangjl6@mails.neu.edu.cn; chuanlin1988@gmail.com; liuy@swc.neu.edu.cn; zhuq@swc.neu.edu.cn;).}
\thanks{Guangjie Han and Shengchao Zhu are with Key Laboratory of Maritime Intelligent Network Information Technology, Ministry of Education, Hohai University. (e-mails: hanguangjie@gmail.com; zhushengchao77@gmail.com.)}

\thanks{The code is available at:\url{https://github.com/PingshengRen0901/SA-MARL}}

\thanks{This is the author-accepted version of
an article published in \emph{IEEE Wireless
Communications}, 2026.
Digital Object Identifier 10.1109/MWC.2026.3742302.}

}

\markboth{Journal of \LaTeX\ Class Files,~Vol.~XX, No.~X, XXX~XXXX}%
{Shell \MakeLowercase{\textit{et al.}}: Bare Demo of IEEEtran.cls for IEEE Journals}
%

\IEEEtitleabstractindextext{%
	\begin{abstract}

With the rapid advancement of underwater networking and multi-agent coordination technologies, autonomous underwater vehicle (AUV) ad-hoc networks have emerged as a pivotal framework for executing complex maritime missions, such as multi-target tracking. 
However, traditional data-centric architectures struggle to maintain operational consistency under highly dynamic topological fluctuations and severely constrained acoustic communication bandwidth. 
This article proposes a scene-adaptive embodied intelligence (EI) architecture for multi-AUV ad-hoc networks, which re-envisions AUVs as embodied entities by integrating perception, decision-making, and physical execution into a unified cognitive loop.
To materialize the functional interaction between these layers, we define a beacon-based communication and control model that treats the communication link as a dynamic constraint-aware channel, effectively bridging the gap between high-level policy inference and decentralized physical actuation.
Specifically, the proposed architecture employs a three-layer functional framework and introduces a Scene-Adaptive MARL (SA-MARL) algorithm featuring a dual-path critic mechanism.
By integrating a scene critic network and a general critic network through a weight-based dynamic fusion process, SA-MARL effectively decouples specialized tracking tasks from global safety constraints, facilitating autonomous policy evolution.
Evaluation results demonstrate that the proposed scheme significantly accelerates policy convergence and achieves superior tracking accuracy compared to mainstream MARL approaches, maintaining robust performance even under intense environmental interference and fluid topological shifts.

		
\end{abstract}
	}
\maketitle
\IEEEdisplaynontitleabstractindextext

\IEEEpeerreviewmaketitle

\section{Introduction}

\label{sec:introduction}

\IEEEPARstart{R}{ecently}, the rapid advancement in underwater communication~\cite{kaushal2016underwater}, multi-agent coordination, and networked control technologies has significantly accelerated ocean exploration and development.
This momentum has fostered the widespread deployment of AUVs~\cite{auvreview} as pivotal enablers for future maritime missions due to their high autonomy and organizational flexibility.
As underwater communication technologies, e.g., underwater acoustic communication technique evolve, the research focus has gradually shifted from single-AUV systems to AUV cluster systems where the AUVs form  self-organizing and distributed intelligence ad-hoc networks, i.e., the multi-AUV ad-hoc networks~\cite{ad-hoc}.
Thus, the AUVs can cooperatively execute complex underwater tasks such as multi-target tracking and encirclement in denied environments. 

As the complexity of underwater missions grows, the paradigm of AUV network management is undergoing a fundamental transformation~\cite{manage}.
Traditional architectures often treat the network as a passive data conduit, which struggle to maintain operational consistency under the highly dynamic topological fluctuations and severely constrained bandwidth inherent to the acoustic communication channel~\cite{acoustic}.

To  overcome the inherent rigidities of passive data-centric architectures, the paradigm of EI~\cite{embodied} offers a compelling framework for re-envisioning underwater network management.
By re-envisioning AUVs as embodied entities rather than mere communication endpoints, this approach integrates perception, autonomous decision-making, and execution into a unified cognitive loop.
Such an integration enables a profound, bi-directional adaptation between high-level mission objectives and the volatile real-time states of the acoustic communication network~\cite{commmunicationnetwork}, effectively mitigating the inconsistencies caused by severe bandwidth constraints and topological fluctuations~\cite{resilientcontrol}.
Crucially, within this Scene-Adaptive EI architecture, MARL functions as the indispensable decision-making core. 
As the "neural engine" of the embodied system, MARL operationalizes the cognitive loop by translating multi-modal environmental perceptions into coordinated physical actions through continuous interactive learning~\cite{auvcommunication}. 
Despite substantial progress, MARL-driven multi-AUV ad-hoc networks still face two critical challenges in the context of intelligent multi-target tracking:

\textbf{Dynamic Scene Adaptability Bottleneck:} Standard MARL algorithms often struggle to balance global mission stability with specialized coordination requirements. In fluid ad-hoc topologies~\cite{ad-hocnet}, gradient conflicts between general safety constraints and scene-specific underwater tasks frequently lead to suboptimal policies or training divergence.

\textbf{Embodied Execution under Constrained Communication:} Achieving precise physical feedback (thrust and torque) from abstract digital policies requires an action-to-control mapping mechanism~\cite{feedback}. In bandwidth-constrained underwater communications~\cite{bandwidth}, maintaining synchronization between decentralized nodes while minimizing control overhead remains a formidable task.

Inspired by the concept of embodied cognitive loops~\cite{embodiedloop}, this paper introduces a Scene-Adaptive EI architecture and a corresponding Scene-Adaptive MARL (SA-MARL) algorithm for autonomous policy learning in multi-AUV ad-hoc networks. 
While Ref.~\cite{beacon} established an SD-MARL architecture and a beacon-based coordination framework for cooperative AUV
target tracking, the present work extends this line by designing a new beacon-based communication and control model and introducing a
scene-adaptive dual-path critic mechanism. The proposed architecture organizes perception, communication, decision-making, and execution into an integrated loop, allowing multi-AUV teams to adapt their cooperative behavior to changing environmental conditions during multi-target tracking.
The main contributions of this article are summarized as follows:

\begin{figure*}[bth]
	\centering
	\includegraphics[width=1.0\linewidth]{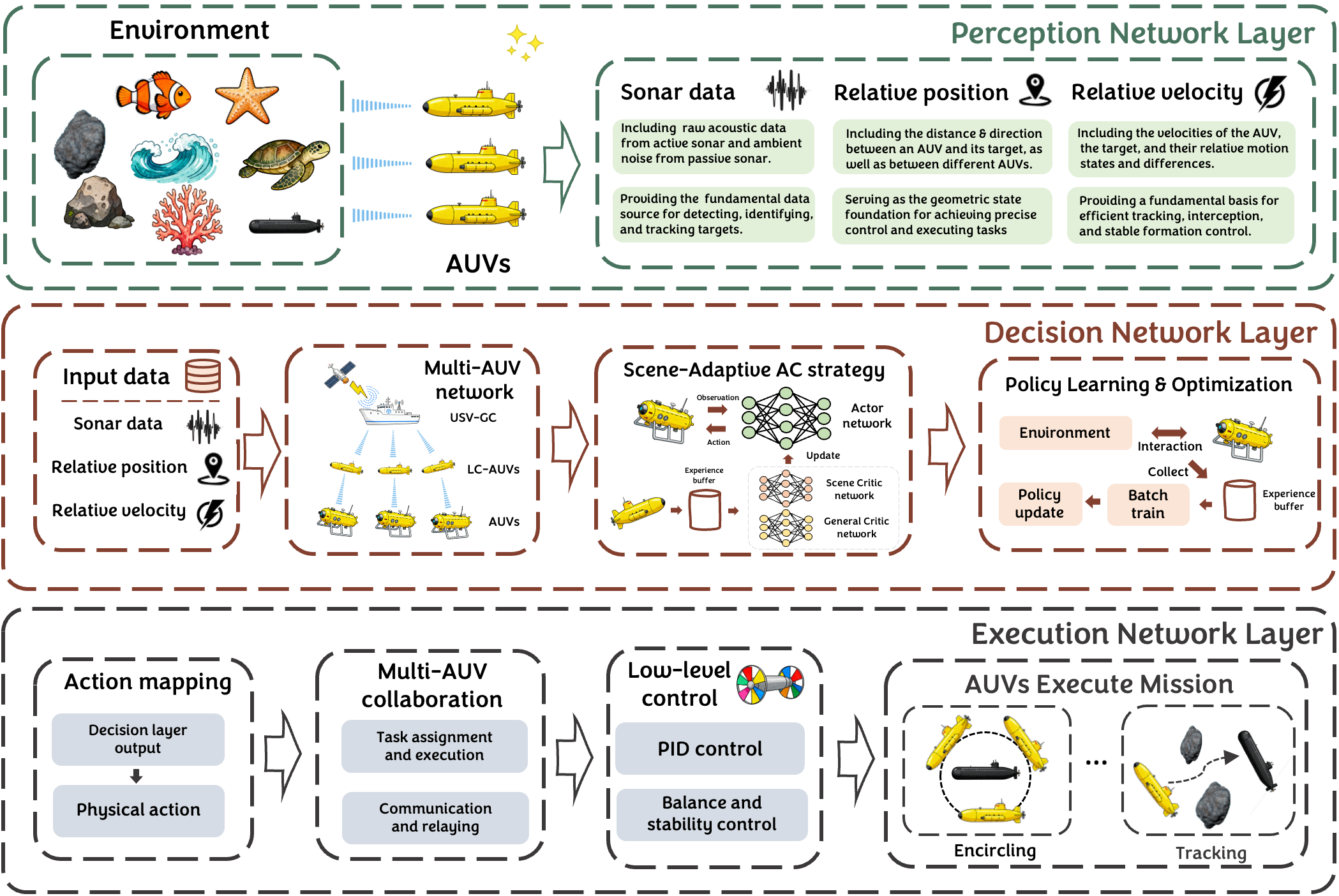}
	\caption{{Proposed Scene-Adaptive EI  architecture for multi-AUV ad-hoc networks}}
	\label{fig1}
\end{figure*}

\begin{itemize}
    \item \textbf{A Scene-Adaptive EI Architecture:} We propose a scene-adaptive three-layer architecture that  ensures that the multi-AUV ad-hoc networks can autonomously shift their coordination strategy to match the real-time requirements of diverse multi-AUV collaborative scenes.
    By introducing a beacon-based communication and control model, this architecture supports cross-layer coordination by encoding cooperative intent and control information into periodic beacon signals, enabling consistent and adaptive coordination in constrained underwater environments.
    
    \item \textbf{The SA-MARL Algorithm with Dual-Path Critics:} We design a SA-MARL algorithm with a dual-path critic mechanism.
    By dynamically integrating a scene critic and a general critic through adaptive weight fusion, the proposed algorithm effectively reconciles conflicts between scene-specific coordination requirements and global task objectives, thereby enabling rapid policy convergence in dynamic ad-hoc topologies.
    
    \item \textbf{A SA-MARL-Driven Smart Tracking Scheme:} We develop a SA-MARL-driven smart tracking scheme to balance precision, formation safety, and motion smoothness. 
    Validated via Unity-based 3D visualization, the scheme maintains superior robustness and tracking accuracy over mainstream MARL approaches, even under intense environmental interference and fluid topological shifts inherent to underwater ad-hoc cluster network.
\end{itemize}

The rest of this paper is organized as follows: Section \ref{Section:2} details the proposed Scene-Adaptive EI architecture. Section \ref{Section:3} describes the proposed SA-MARL algorithm. Section \ref{Section:4} introduces the smart tracking scheme based on the proposed SA-MARL. Finally, Section \ref{Section:5} presents comprehensive simulation results and corresponding 3D visualizations via the Unity engine.

\section{Proposed Scene-Adaptive EI Architecture For Multi-AUV ad-hoc Networks}\label{Section:2}

To address the challenges posed by dynamic topology changes and constrained communication in multi-AUV ad-hoc networks, we propose a scene-adaptive embodied-intelligence (EI) architecture, as shown in Fig.~\ref{fig1}.

The EI architecture defines how information and control signals circulate through the multi-AUV system.
The perception layer produces local observations, the beacon-based communication model exchanges coordination information among network nodes, the decision layer generates cooperative actions using the proposed SA-MARL method, and the execution layer maps these actions to low-level vehicle commands. 

Within this architecture, SA-MARL implements the decision function of the decision network layer. Therefore, EI specifies the organization and interfaces of the complete embodied system, whereas SA-MARL specifies the learning mechanism used to select cooperative actions.

\subsection{Perception Network Layer}
Acting as the foundation of the architecture, this layer is responsible for capturing raw signals from the dynamic marine environment and transforming them into high-level observational states required for intelligent decision-making. 
Specifically designed for the multi-AUV ad-hoc networks, it functions as a distributed sensory system that ensures comprehensive situational awareness across shifting network topologies. By integrating active and passive sonar data, the layer filters ambient noise to provide a continuous stream for target detection and identification within the ad-hoc networks. Furthermore, it constructs a precise geometric foundation for node coordination by calculating real-time relative distances and bearings between AUVs and targets. By monitoring ego-motion and relative velocity, this layer provides the essential dynamic basis for formulating tracking paths and maintaining stable formations despite the decentralized, self-organizing nature of the ad-hoc networks.

\subsection{Decision Network Layer}
Positioned as the intelligent control center, this layer employs MARL to enable adaptive policy evolution in dynamic marine environments. 
The architecture assigns global coordination to an Unmanned Surface Vessel-based Global Controller (USV-GC), while the Local Control AUV (LC-AUV) performs regional task allocation and centralized policy learning.
The decision layer implements SA-MARL within a centralized-training-with-decentralized-execution (CTDE) framework. 
During training, the LC-AUV uses joint cluster information and beacon feedback to train the actor and critic networks.
During deployment, each AUV makes autonomous decisions from its local observations and received coordination information. The resulting policy decisions are passed to the execution layer for
vehicle-level control.

\subsection{Execution Network Layer}
Serving as the terminal stage of the system’s closed loop, the execution network layer facilitates the embodied transition from digital policy inference to concrete physical feedback. 
Within the dynamic framework of the multi-AUV ad-hoc networks, this layer employs an action mapping mechanism to translate abstract decision vectors into thrust and torque commands for each AUV's propulsion system. 
To maintain collective synergy, it further integrates low-level PID control and stability modules that counter unpredictable underwater disturbances, ensuring the cluster strictly adheres to its intended formation and trajectory.
This high-level integration across the ad-hoc nodes ultimately drives the fulfillment of complex missions, such as target encirclement and tracking, thereby completing the comprehensive circuit from environmental perception to physical task execution.

\begin{figure}[bth]
	\centering
	\includegraphics[width=1.0\linewidth]{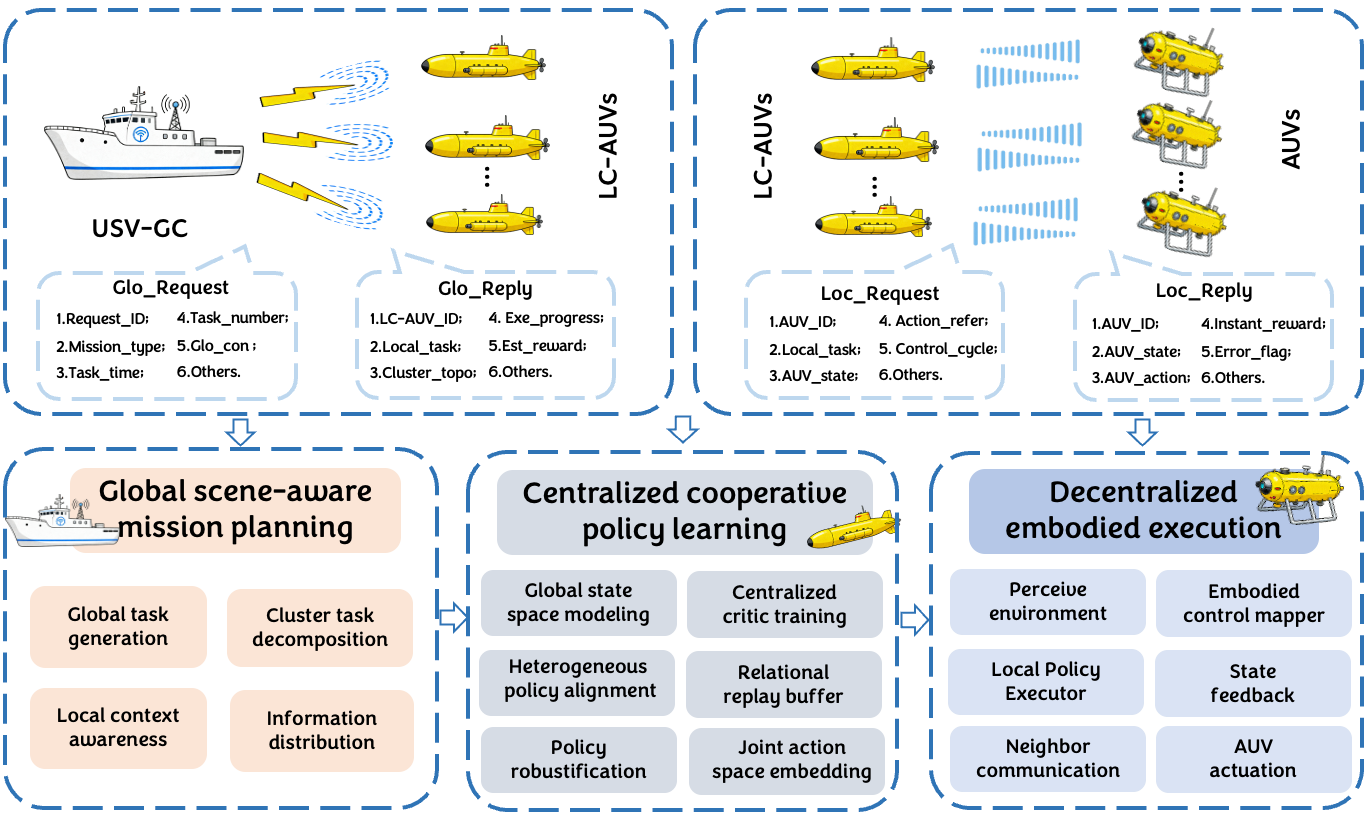}
	\caption{Beacon-based  communication and control model in multi-AUV ad-hoc networks}
	\label{fig3}
\end{figure}
\subsection{Beacon-Based  Communication And Control Model}\label{Section:6-1}

\begin{figure*}[bth]
	\centering
	\includegraphics[width=1.0\linewidth]{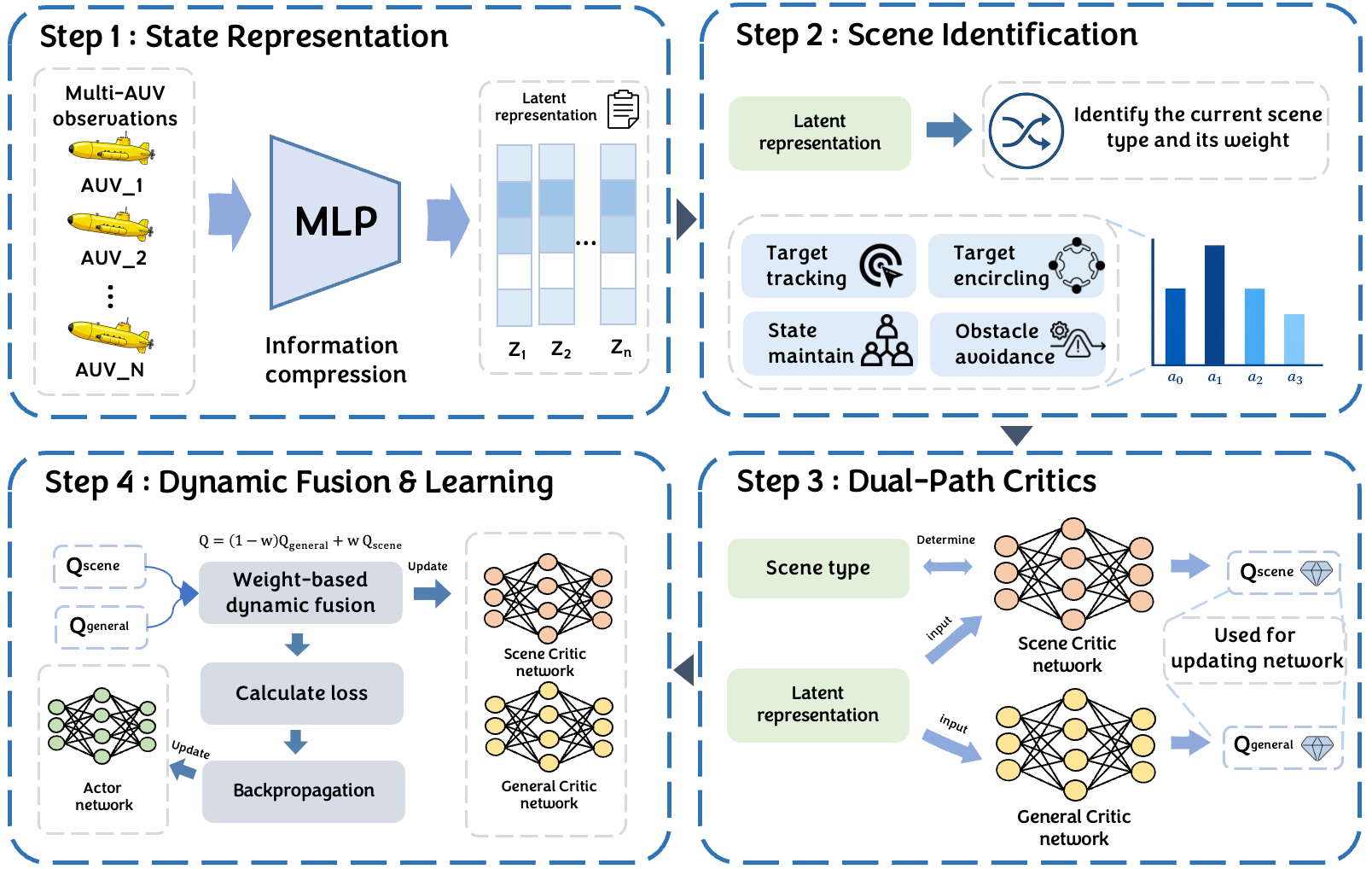}
	\caption{Proposed Scene-Adaptive MARL algorithm}
	\label{fig2}
\end{figure*}


To materialize the functional interaction between the aforementioned layers, we further define a beacon-based communication and control model (as shown in Fig.~\ref{fig3}). 
This model transcends simple data exchange by treating the communication link as a dynamic constraint-aware channel, encoding cooperative intent, synchronization sequences, and control gradients into periodic beaconing.

Specifically, the core of this model lies in "Underwater Communication-Driven Control Modulation": it utilizes discrete beacon signals to calibrate the continuous control loops of the AUVs, ensuring collective synchronization despite the low-bandwidth and high-latency nature of underwater ad-hoc networks. 
By referring to our previous work~\cite{beacon}, we design the beacon mechanism to achieve robust and efficient coordination under constrained communication conditions.
The beacon messages are transmitted periodically and carry mission-, task-, state-, progress-, reward-, and execution-feedback information. Low-level
control commands are generated locally by each AUV rather than continuously
exchanged through the acoustic channel, thereby limiting the communication
overhead of decentralized execution.
In the communication-stress tests, beacon loss and one-way communication delay are introduced to examine their effects on decentralized coordination. When a beacon is lost or delayed, the receiving AUV retains the most recently
received valid beacon until a newer beacon becomes available.
The control and interaction logic is detailed as follows:

\textbf{Global Mission Planning (USV-GC $\leftrightarrow$ LC-AUV):} 
The Unmanned Surface Vessel Global Controller (USV-GC) initiates the hierarchy by issuing Glo\_Request beacons, which encapsulate the Mission\_type, Glo\_con (global constraints), and Task\_number required to coordinate regional clusters.
In return, the LC-AUV provides Glo\_Reply beacons reporting Cluster\_topo, Exe\_progress, and Est\_reward, allowing the USV-GC to maintain global situational awareness and perform high-level mission decomposition.

\textbf{Centralized Cooperative Policy Learning (LC-AUV $\leftrightarrow$ AUV):} 
Serving as the regional intelligence hub, the LC-AUV manages its cluster using Loc\_Request beacons to distribute Local\_task details and Control\_cycle parameters.
Rather than being passive data nodes, the follower AUVs function as active embodied participants; they respond with Loc\_Reply beacons that feed back AUV\_state, Instant\_reward, and execution Error\_flag. These heterogeneous data streams are integrated into a relational replay buffer, enabling the LC-AUV to perform centralized critic training and construct a joint action space embedding that accounts for inter-AUV dependencies.

\textbf{Decentralized Embodied Execution:} 
This stage translates the learned cooperative policies into autonomous physical actions.
Individual AUVs utilize an embodied control mapper to convert local policy outputs into direct actuation commands, enabling the AUVs to perceive their environment and communicate with neighbors to maintain formation. 
This decentralized approach ensures that the AUV cluster can execute missions such as target tracking or encirclement with high robustness, even during abrupt shifts in the ad-hoc network topology.

\section{Proposed Scene-Adaptive MARL Algorithm}\label{Section:3}

Based on the Scene-Adaptive EI architecture, this section proposes the Scene-Adaptive MARL algorithm (as shown in Fig.~\ref{fig2}) as the core of the decision network layer to facilitate autonomous policy evolution within the dynamic multi-AUV ad-hoc networks.
To address the diverse and shifting requirements of multi-AUV coordination scenarios, SA-MARL utilizes a specialized four-step pipeline designed to maintain policy alignment with the real-time operational scene:

\textbf{Step 1 - State Representation:}
High-dimensional multi-AUV observations from the ad-hoc networks are aggregated and processed through a Multi-Layer Perceptron (MLP). 
This compression stage extracts a latent representation ($Z_1, Z_2, \dots, Z_n$), optimizing data efficiency for bandwidth-constrained underwater communication while capturing the essential features of the AUVs' operational state.

\textbf{Step 2 - Scene Identification:}
Utilizing the latent representation, the system autonomously identifies the current collaborative scene such as target tracking, encirclement, or obstacle avoidance. 
It assigns scene-relevance scores $(a_0,a_1,a_2,a_3)$ to the candidate coordination scenes.
These scores are generated from the latent representation by a trainable scene-identification module and updated during SA-MARL training.
By evaluating the magnitude of these weights in real-time, the system selects the dominant environmental scene to prioritize the corresponding joint coordination strategy. 
This weight-based selection ensures that the AUV cluster adaptively shifts its operational focus and maintains collective synergy as the mission environment evolves.

\textbf{Step 3 - Dual-Path Critics:}
To simultaneously assess different aspects of the cluster's performance, this algorithm employs a dual-stream evaluation mechanism.
A scene critic network calculates the scene-specific value based on the dominant scene type selected in the previous stage, while a general critic network ensures overall system integrity by focusing on stability and operational safety.
By maintaining these parallel evaluation paths, the system ensures that specialized coordination tasks do not compromise the global safety constraints of the multi-AUV ad-hoc networks.

\textbf{Step 4 - Dynamic Fusion and Learning:}

The outputs of the general critic network and the scene critic network are adaptively integrated to obtain a unified action-value estimate:
\begin{equation}
Q_t=(1-w_t)Q_{\mathrm{general},t}+w_tQ_{\mathrm{scene},t},
\label{eq:critic_fusion}
\end{equation}
where $Q_{\mathrm{general},t}$ and $Q_{\mathrm{scene},t}$ are the value estimates produced by the general critic network and the scene critic network, respectively.
The fusion weight is computed as $w_t=\sigma(g_{\phi}(z_t))\in[0,1]$, where $z_t$ represents the current coordination scene, $g_{\phi}(\cdot)$ is a learnable gating function parameterized by $\phi$, and $\sigma(\cdot)$ is the sigmoid function. Consequently, the contributions of the two critics are dynamically adjusted according to the current scene, balancing general coordination objectives with scene-specific requirements.

\section{Proposed Smart Tracking Scheme Based on SA-MARL}\label{Section:4}


In this section, multi-target tracking is formulated as a multi-objective learning problem within the SA-MARL framework. The reward function incorporates four complementary objectives: target tracking, formation spacing, motion smoothness, and velocity consistency. For AUV \(i\) at time step \(t\), the reward is defined as
\begin{equation}
r_i^t =
0.8r_{i,\mathrm{tr}}^t
+0.4r_{i,\mathrm{sp}}^t
+0.1r_{i,\mathrm{sm}}^t
+1.0r_{i,\mathrm{vc}}^t.
\label{eq:total_reward}
\end{equation}
where \(r_{i,\mathrm{tr}}^t\), \(r_{i,\mathrm{sp}}^t\), and \(r_{i,\mathrm{sm}}^t\) denote the target-tracking, formation-spacing, and motion-smoothness terms for AUV \(i\), respectively. The term \(r_{i,\mathrm{vc}}^t\) denotes a team-level velocity-consistency term shared by the AUVs within the cluster. The coefficients in (\ref{eq:total_reward}) were selected through a grid search to balance the four objectives and fixed for all experiments.

The target-tracking term penalizes the deviation between the current AUV--target distance and the desired tracking distance. The formation-spacing term penalizes deviations from the prescribed inter-AUV spacing to maintain the desired formation geometry. The motion-smoothness term penalizes velocity variations exceeding a predefined threshold, thereby reducing abrupt changes in motion. The velocity-consistency term penalizes the aggregate mismatch between the velocities of the AUVs and their assigned targets, encouraging coordinated tracking at the cluster level. Thus, the target-tracking and velocity-consistency terms directly address tracking and coordination, while the formation-spacing and motion-smoothness terms regulate formation geometry and motion behavior. Their weighted sum in (\ref{eq:total_reward}) provides the learning signal for policy optimization.

\section{Evaluations}\label{Section:5}
This section presents comprehensive comparative evaluations between the proposed approach and several mainstream MARL-driven approaches. 

\subsection{Simulation setup}\label{Section:7-1}
All the simulations are conducted on a MacBook Air (2024) equipped with an Apple M3 processor and 16 GB memory. In the simulation, both the AUV and the target entities are abstracted as dynamic particles evolving within a three-dimensional continuous space.
To represent constrained underwater acoustic links, the maximum communication range is set to 300~m, the nominal data rate is 9.6~kbps, and beacons are transmitted every 0.5~s. 
Communication impairments are evaluated by varying the packet-loss
rate from 0\% to 40\% and the one-way delay from 0.2 to 1.0~s.

\subsection{Results}\label{Section:6-2}
To evaluate the effectiveness of the proposed SA-MARL-driven scheme for
multi-target tracking, we conduct a comprehensive comparison with a set of mainstream MARL methods. Specifically, DSBM and MA-A3C are included as representative multi-agent benchmark methods, while MAAC, MADDPG, MAPPO, MASAC, and MATD3 represent widely used multi-agent actor--critic and policy-optimization approaches~\cite{marlsurvey}. These methods are compared with SA-MARL in terms of convergence speed and tracking accuracy.

\begin{figure*}
\centering

\subfloat[4 AUVs tracking 2 targets ]{%
  \includegraphics[width=0.32\textwidth]{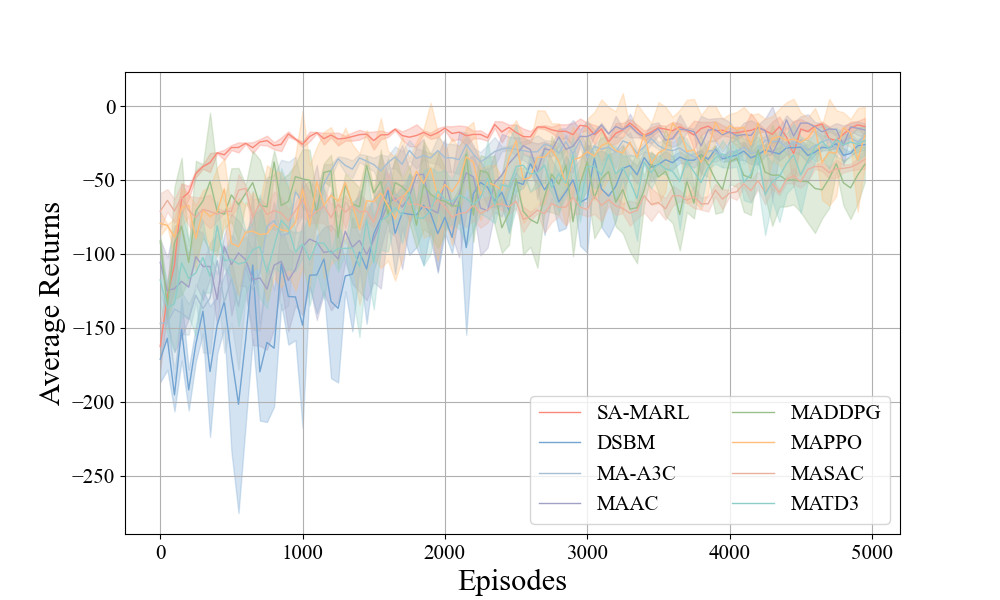}%
  \label{fig:4a}%
}\hfill
\subfloat[6 AUVs tracking 3 targets ]{%
  \includegraphics[width=0.32\textwidth]{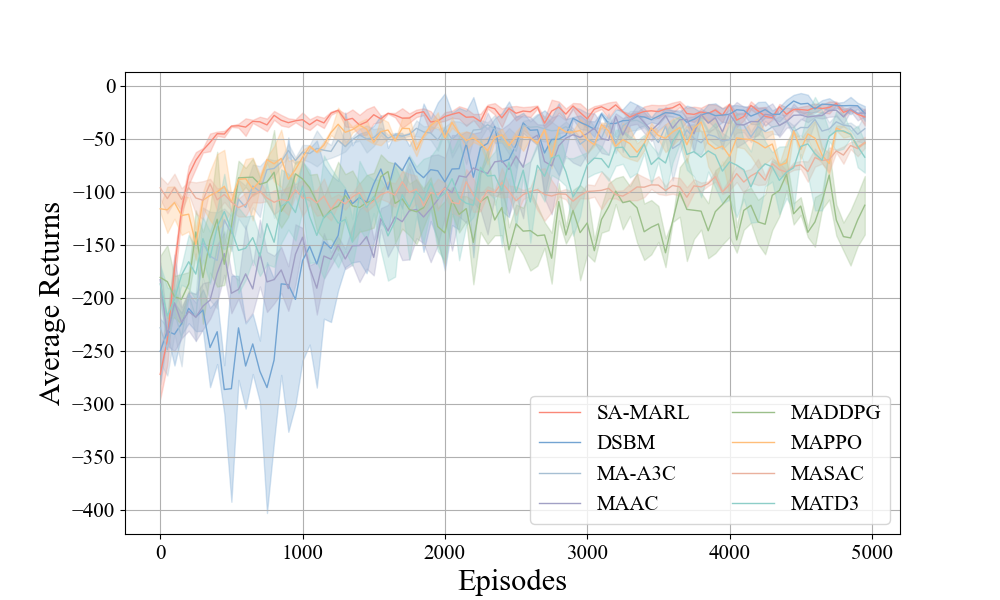}%
  \label{fig:4b}%
}\hfill
\subfloat[12 AUVs tracking 4 targets ]{%
  \includegraphics[width=0.32\textwidth]{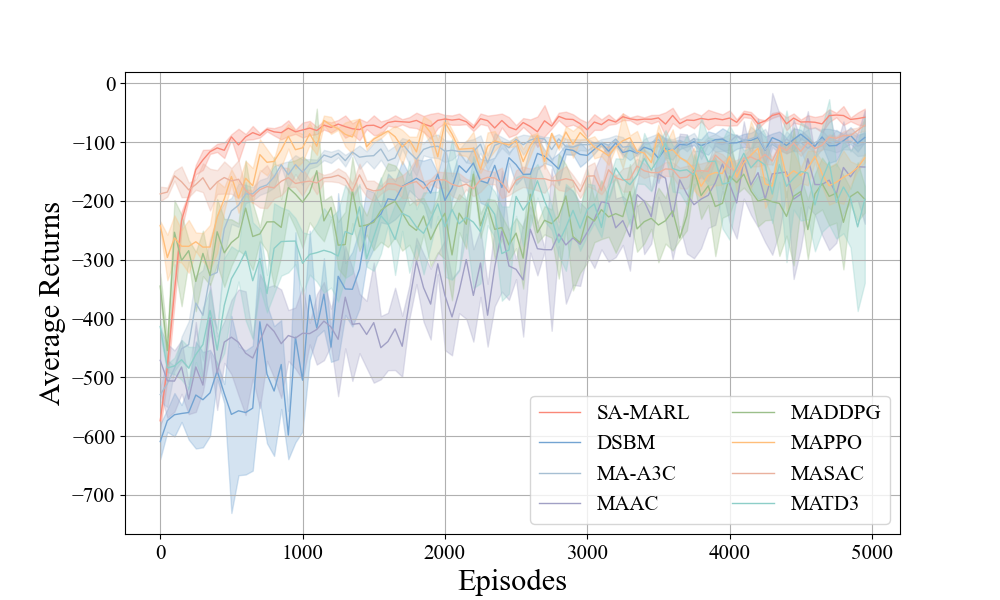}%
  \label{fig:4c}%
}

\subfloat[4 AUVs tracking 2 targets (interfered)]{%
  \includegraphics[width=0.32\textwidth]{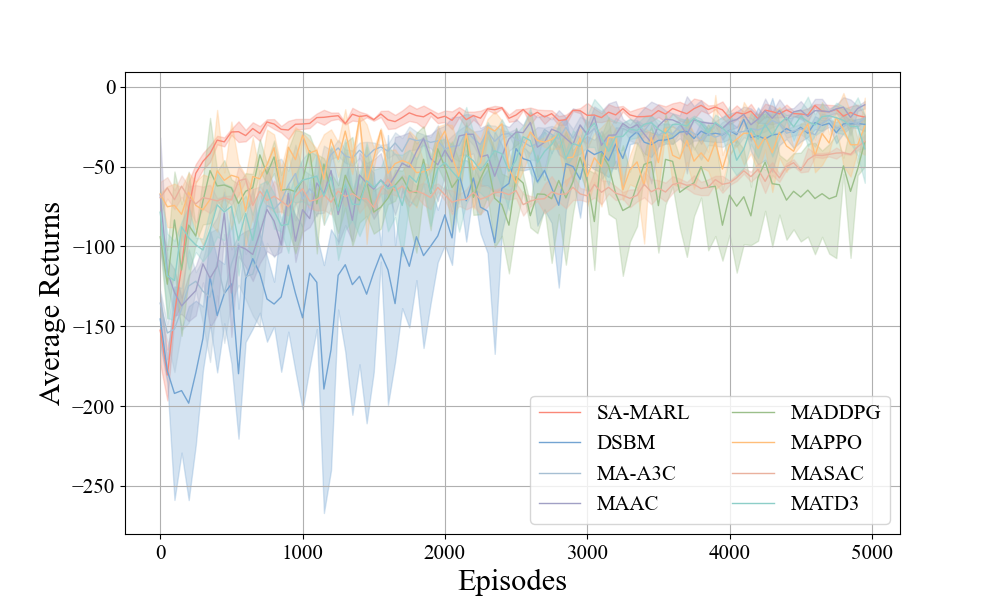}%
  \label{fig:4d}%
}\hfill
\subfloat[6 AUVs tracking 3 targets (interfered)]{%
  \includegraphics[width=0.32\textwidth]{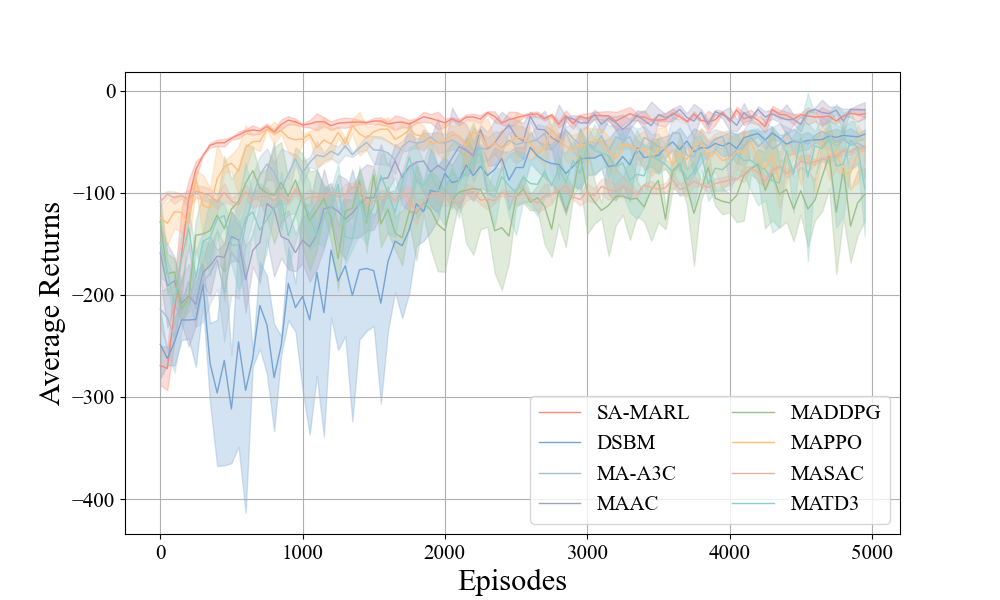}%
  \label{fig:4e}%
}\hfill
\subfloat[12 AUVs tracking 4 targets (interfered)]{%
  \includegraphics[width=0.32\textwidth]{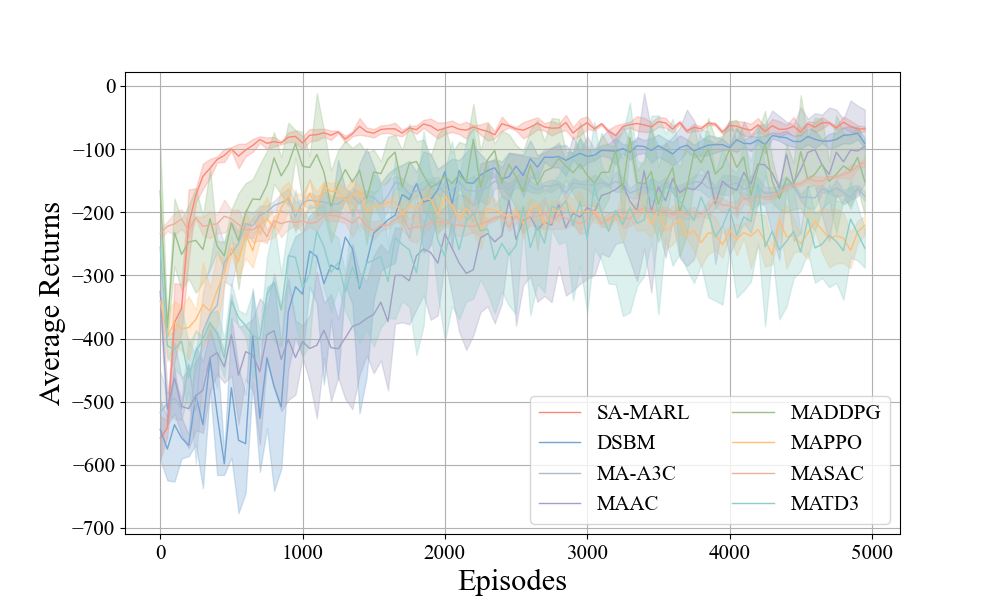}%
  \label{fig:4f}%
}

\caption{Convergence speed comparison}
\label{fig:4}

\end{figure*}

\makeatletter
\def\blfootnote{\gdef\@thefnmark{}\@footnotetext}
\makeatother


\begin{figure*}[bth]
\centering

\subfloat[Tracking accuracy under ideal communication environment]{%
    \includegraphics[width=0.40\textwidth]{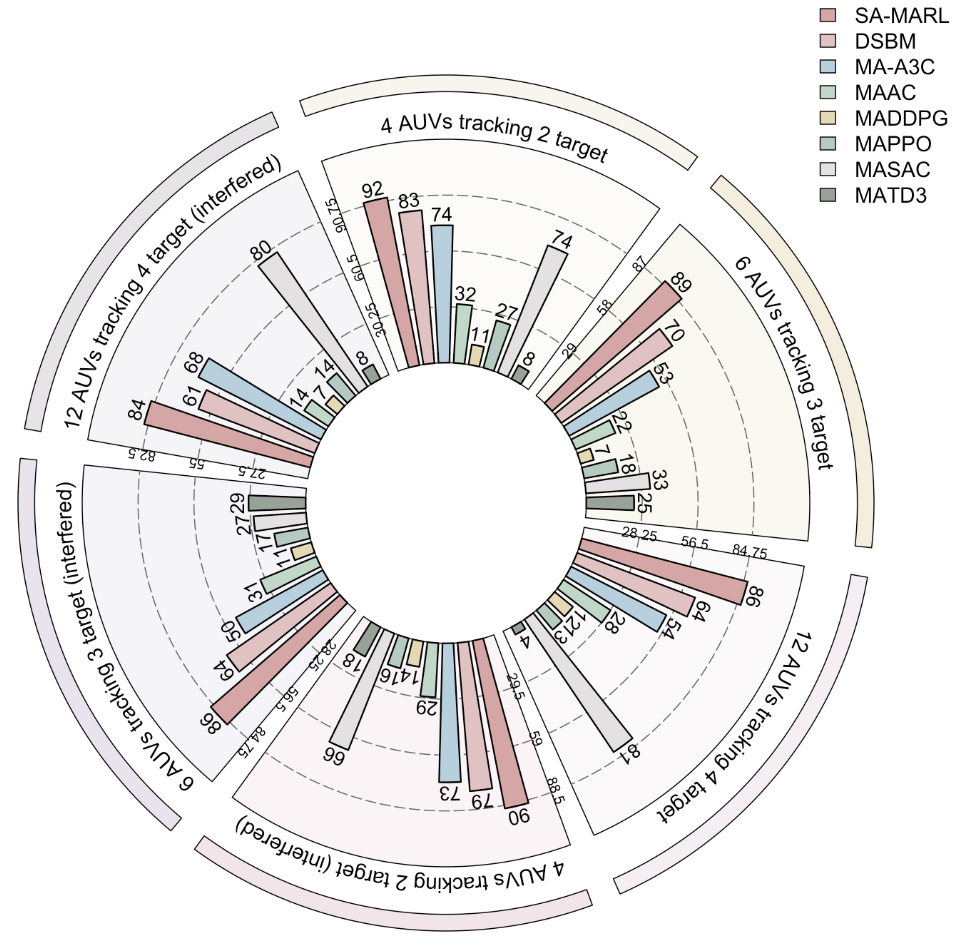}%
    \label{fig:5a}%
}\hfill
\subfloat[Tracking accuracy under different communication environment]{%
    \includegraphics[width=0.59\textwidth]{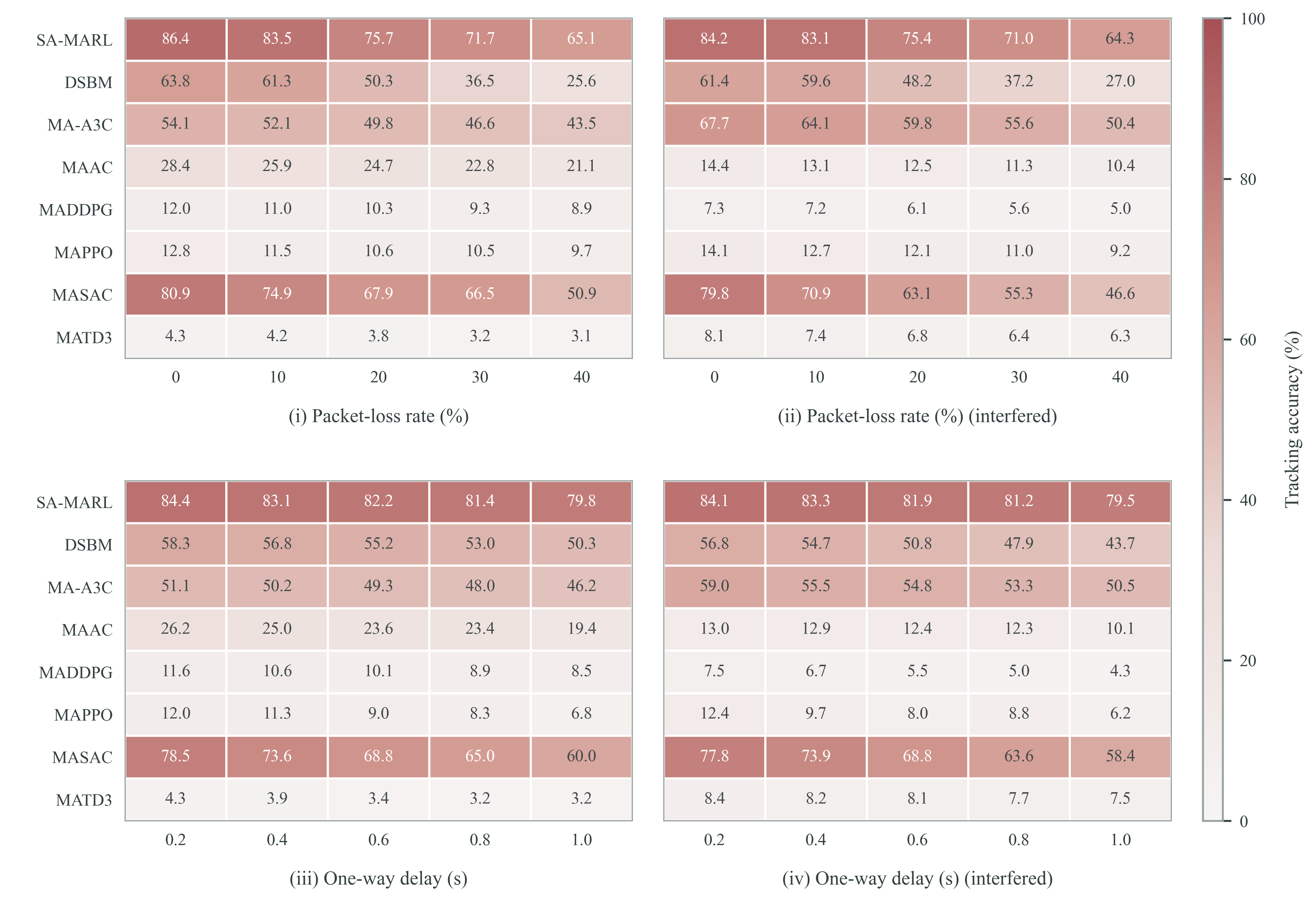}%
    \label{fig:5b}%
}

\caption{Tracking accuracy comparison.}
\label{fig:5}
\end{figure*}

\begin{figure*}
\centering

\subfloat[Initial phase in 4 AUVs tracking 2 targets]{%
  \includegraphics[width=0.33\textwidth]{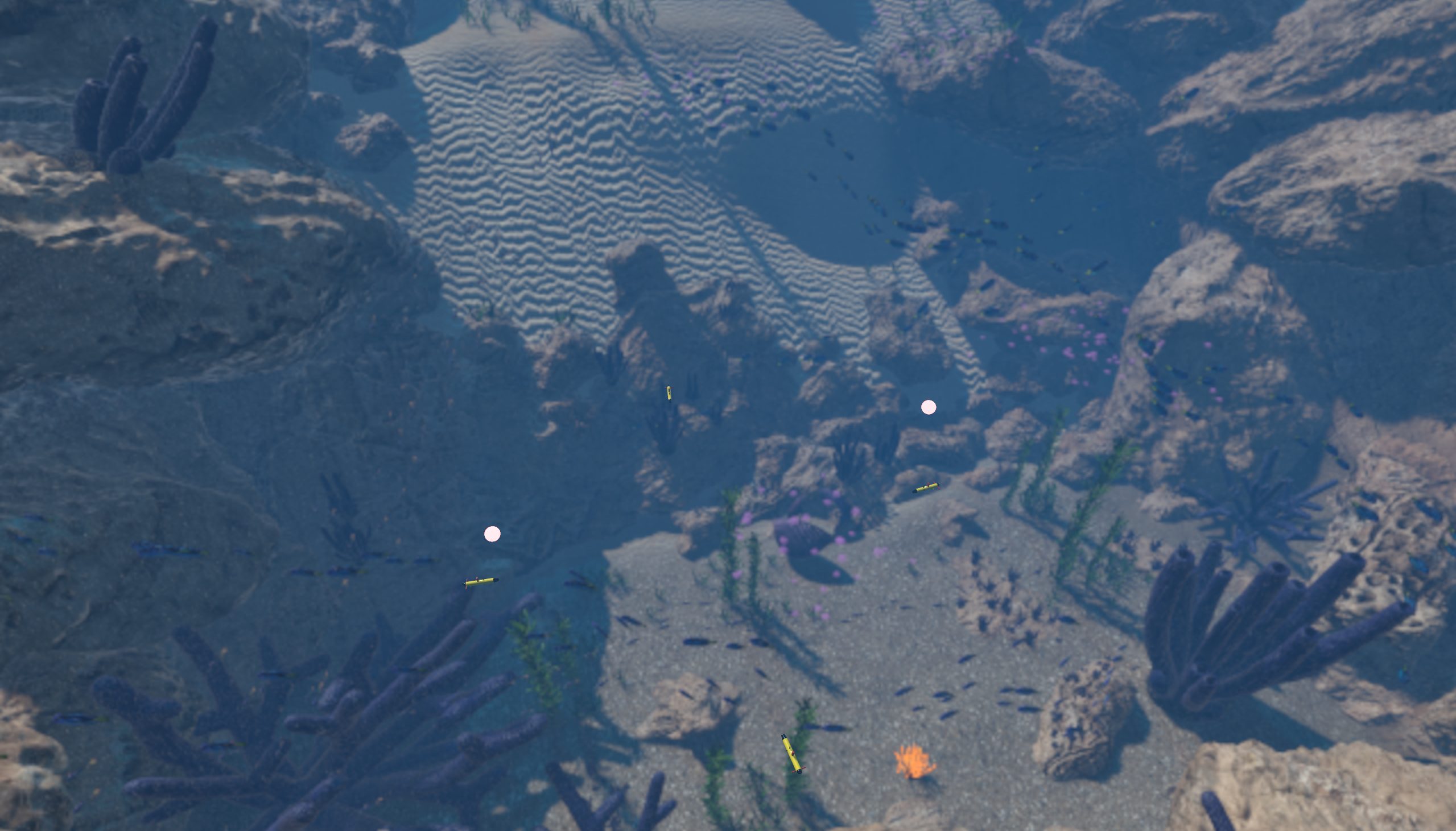}%
  \label{fig:6a}%
}\hfill
\subfloat[Mid-phase in 4 AUVs tracking 2 targets]{%
  \includegraphics[width=0.33\textwidth]{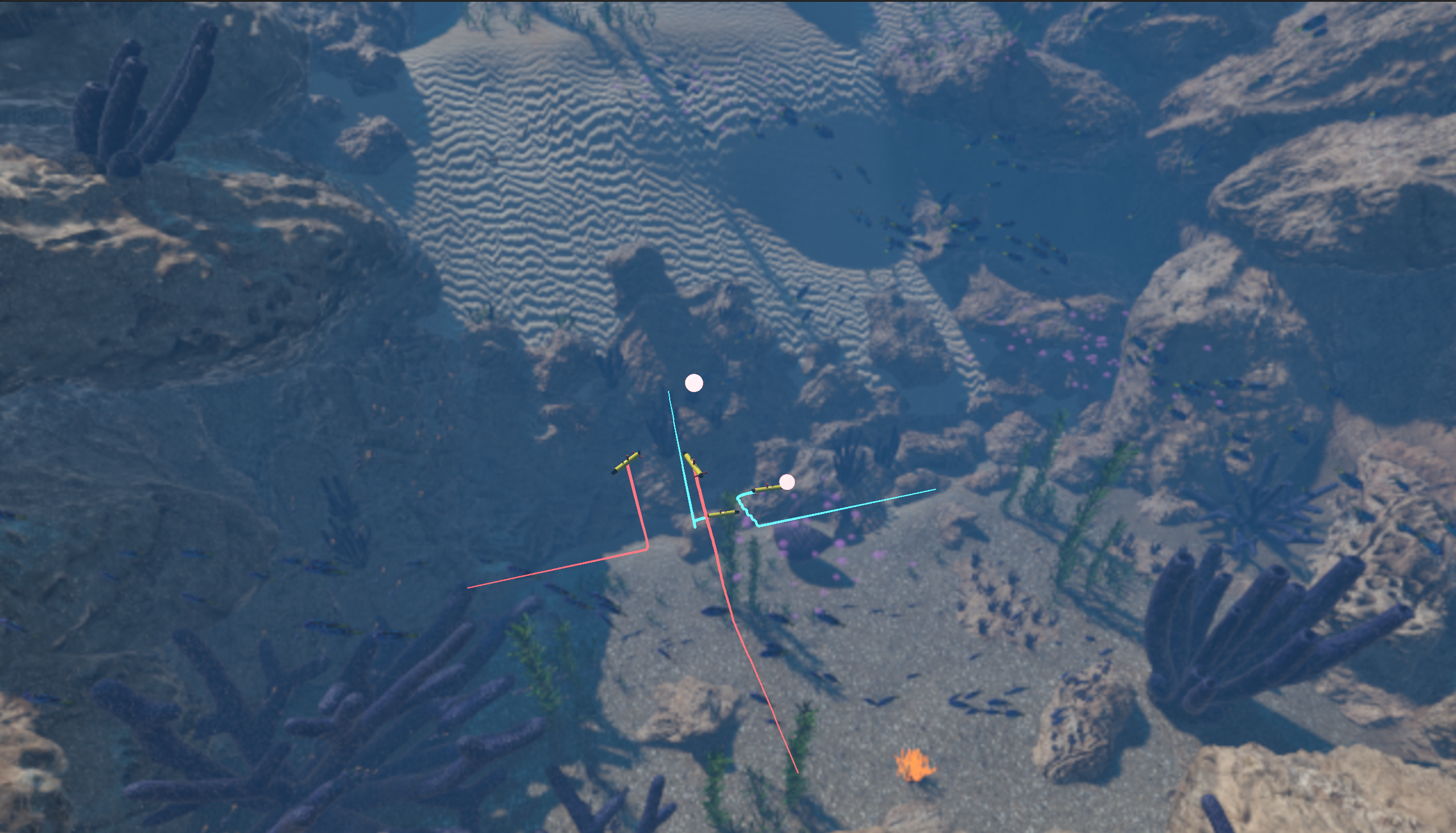}%
  \label{fig:6b}%
}\hfill
\subfloat[Final phase in 4 AUVs tracking 2 targets]{%
  \includegraphics[width=0.33\textwidth]{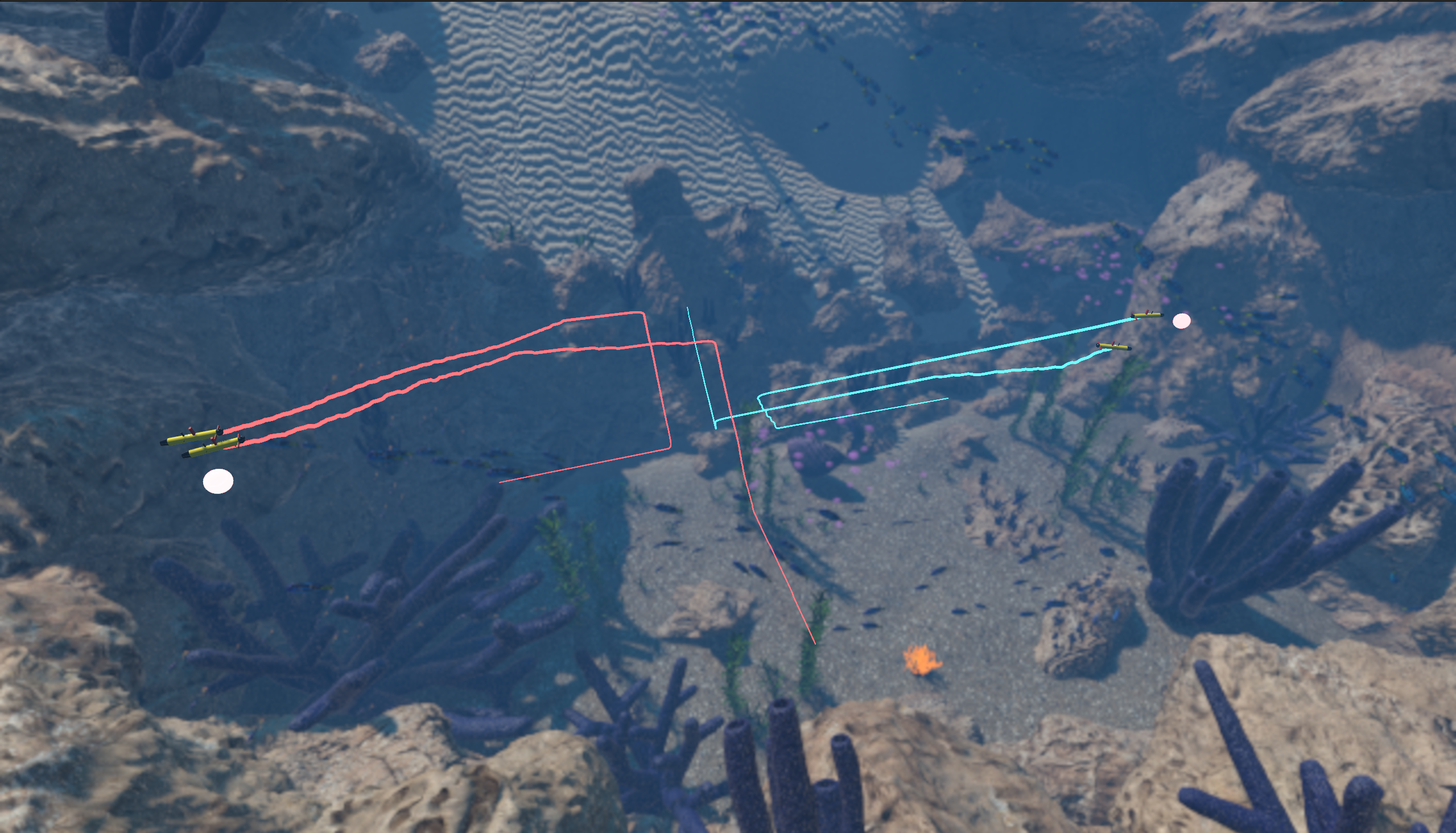}%
  \label{fig:6c}%
}

\subfloat[Initial phase in 6 AUVs tracking 3 targets]{%
  \includegraphics[width=0.33\textwidth]{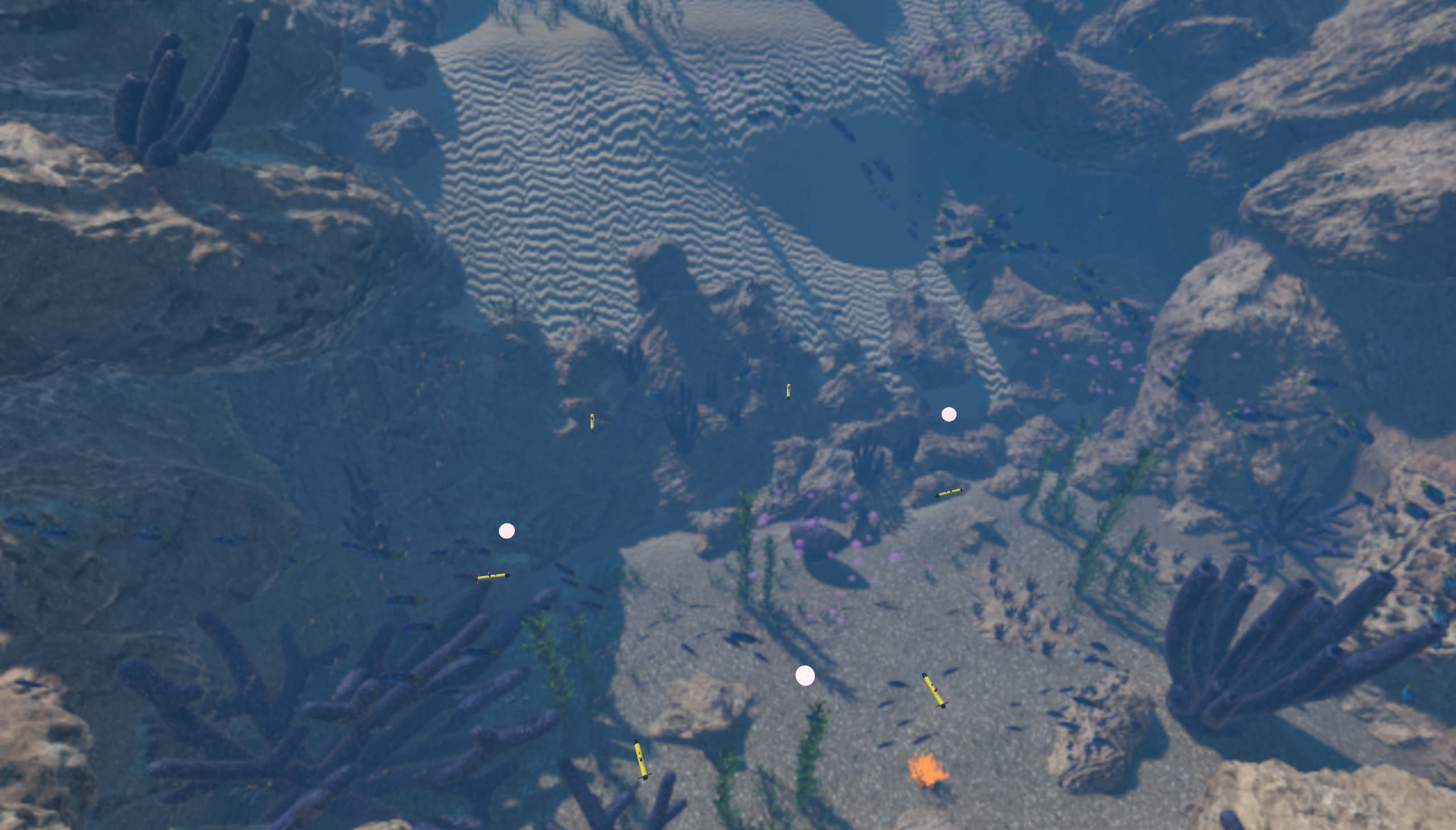}%
  \label{fig:6d}%
}\hfill
\subfloat[Mid-phase in 6 AUVs tracking 3 targets]{%
  \includegraphics[width=0.33\textwidth]{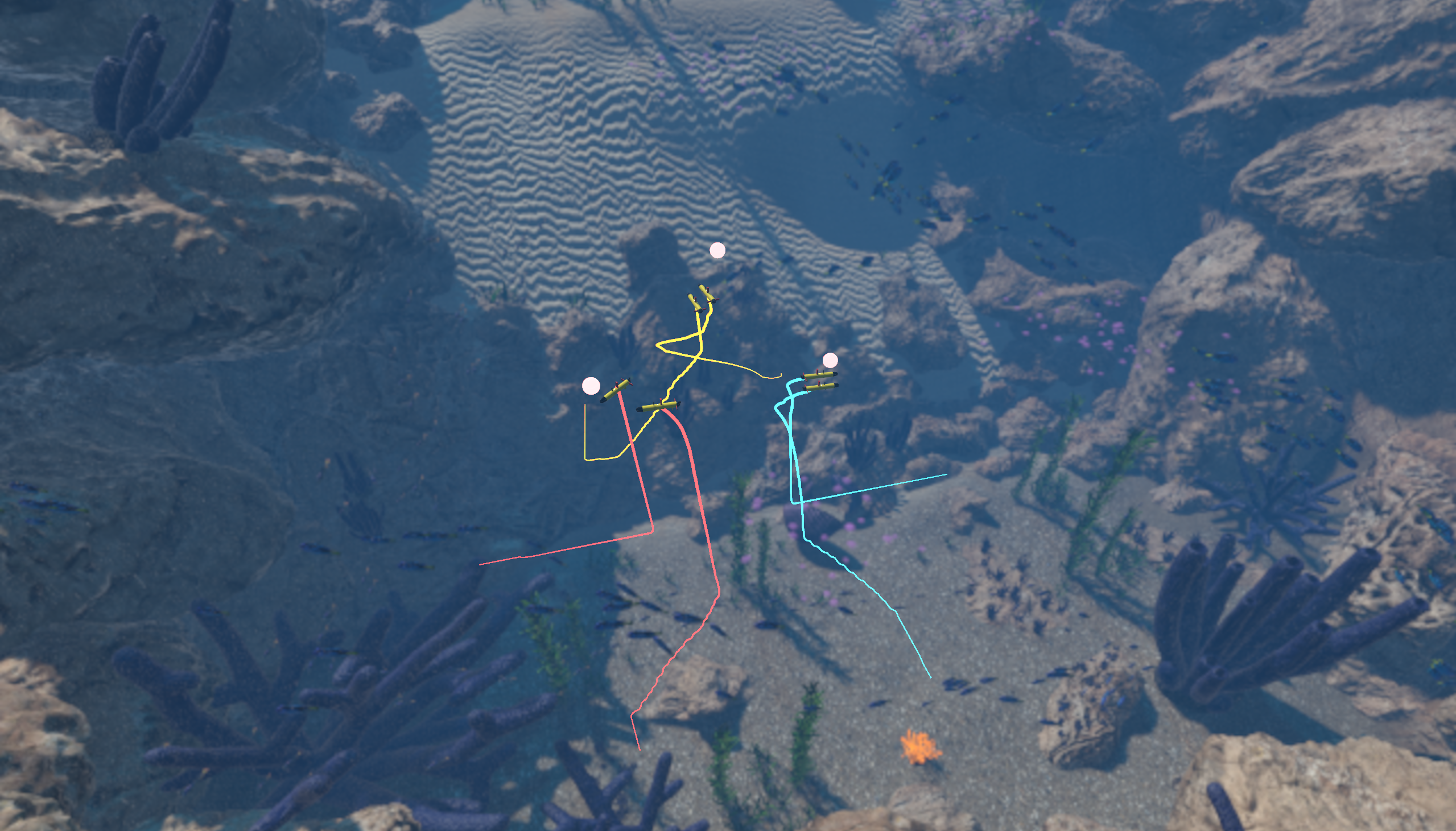}%
  \label{fig:6e}%
}\hfill
\subfloat[Final phase in 6 AUVs tracking 3 targets]{%
  \includegraphics[width=0.33\textwidth]{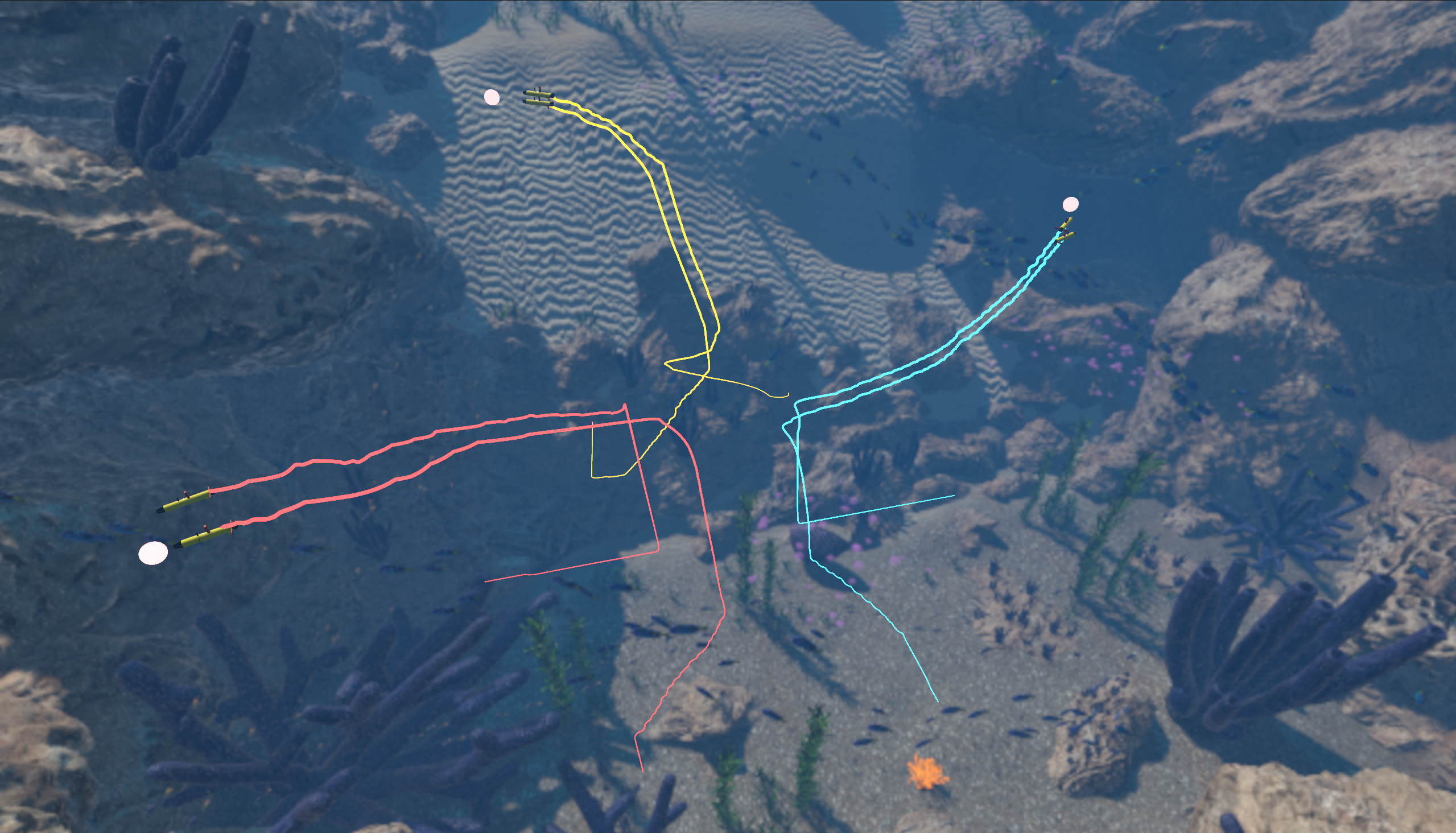}%
  \label{fig:6f}%
}

\subfloat[Initial phase in 12 AUVs tracking 4 targets]{%
  \includegraphics[width=0.33\textwidth]{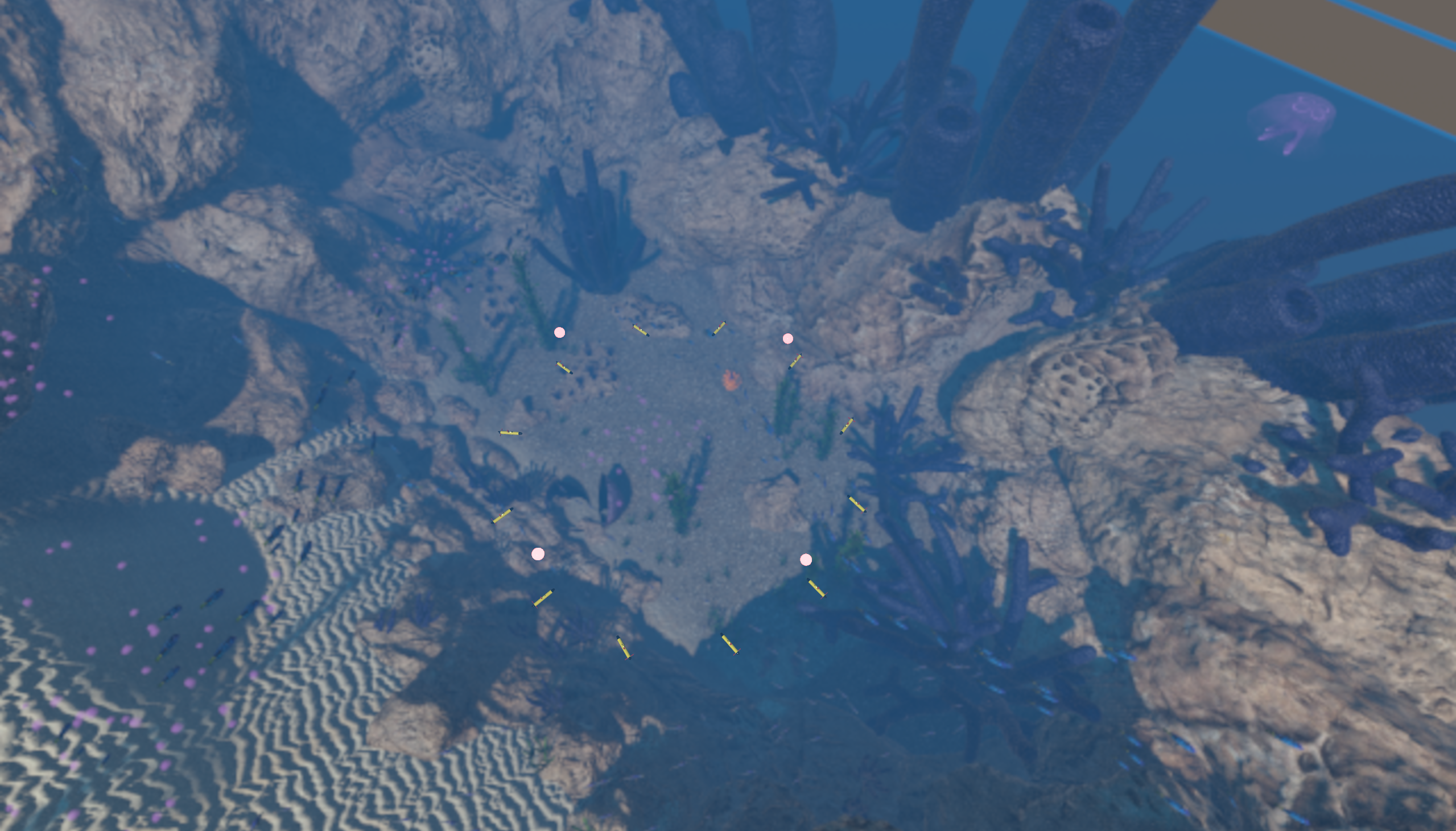}%
  \label{fig:6g}%
}\hfill
\subfloat[Mid-phase in 12 AUVs tracking 4 targets]{%
  \includegraphics[width=0.33\textwidth]{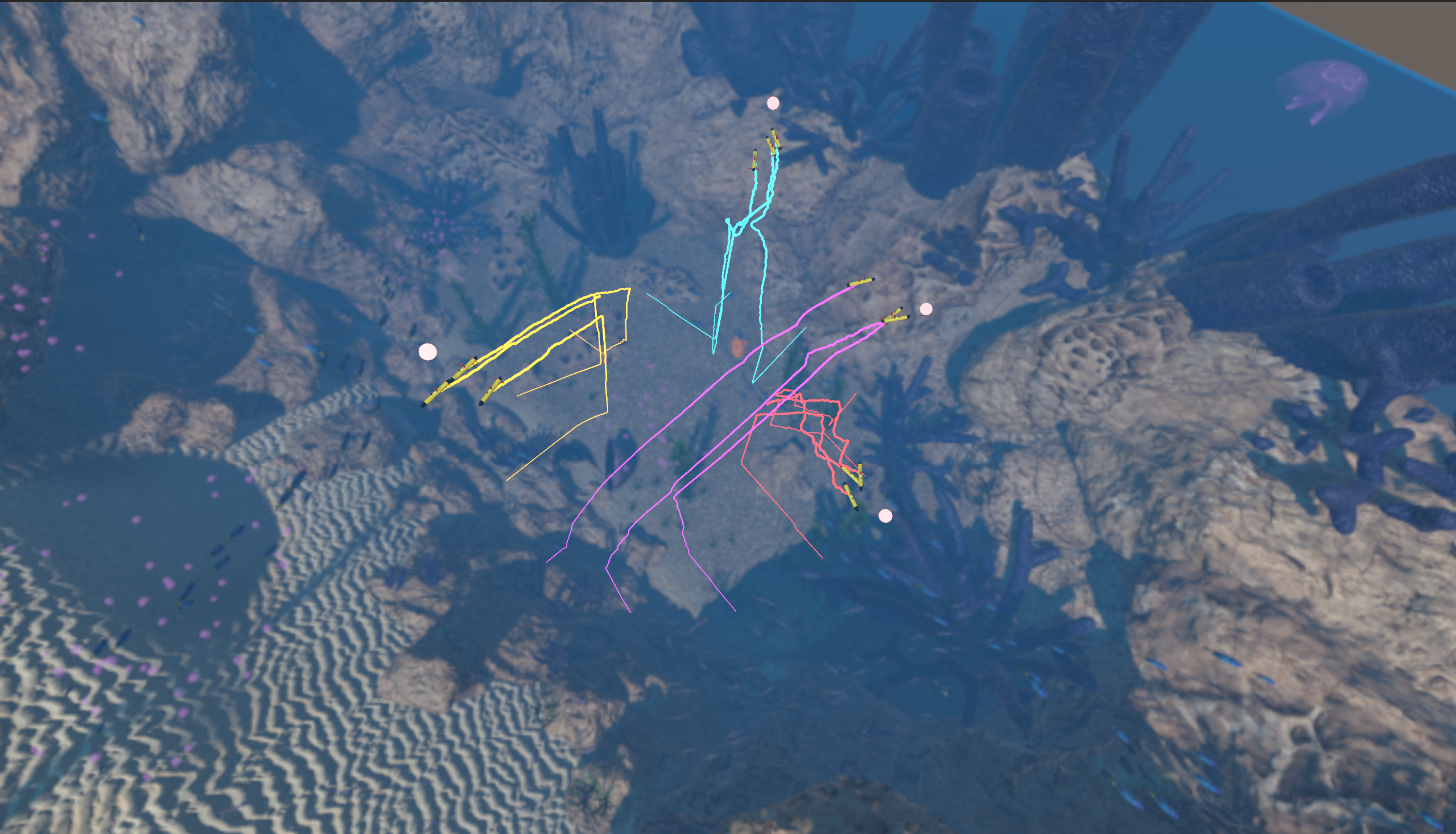}%
  \label{fig:6h}%
}\hfill
\subfloat[Final phase in 12 AUVs tracking 4 targets]{%
  \includegraphics[width=0.33\textwidth]{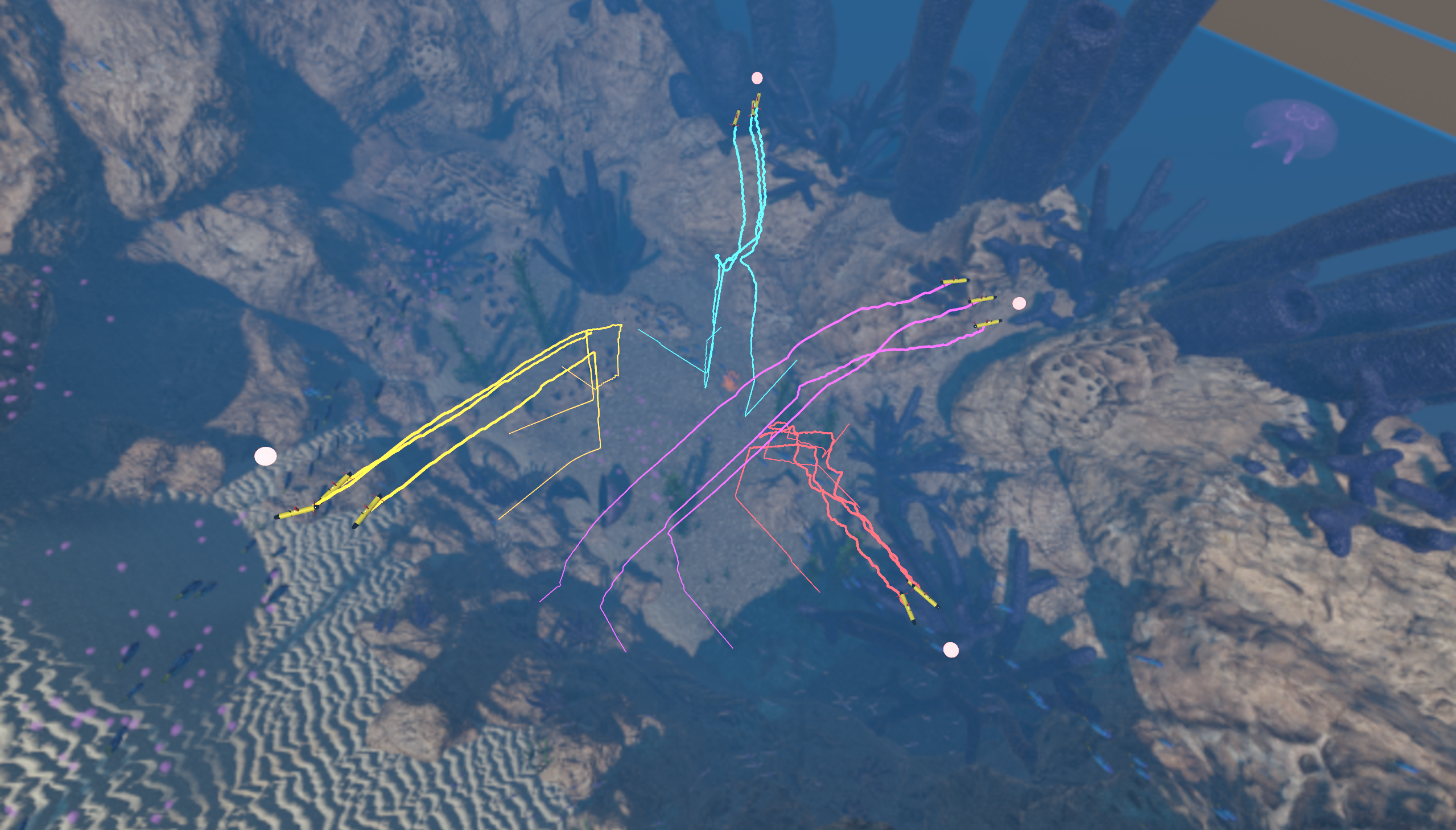}%
  \label{fig:6i}%
}
\caption{Unity-based 3D visualization of the  multi-target tracking process}
\label{fig:6}

\end{figure*}

We first assess the learning efficiency of our proposed approach by comparing its convergence performance against the aforementioned MARL-driven methods.
As illustrated in Fig. \ref{fig:4a}--\ref{fig:4c}, comprehensive evaluations are conducted across three mission scales: 4 AUVs tracking 2 targets, 6 AUVs tracking 3 targets, and 12 AUVs tracking 4 targets.
The experimental results demonstrate that the SA-MARL-driven approach (represented by the red curve) consistently achieves significantly higher average returns and exhibits a faster convergence rate compared to all other approaches across every scale.

Unlike MADDPG, MAPPO, MASAC, and MATD3, SA-MARL explicitly combines scene identification with two functionally distinct critics. The Scene Critic emphasizes scene-specific tracking requirements, whereas the General Critic preserves general coordination objectives.
Although MATD3 also employs twin critics, both critics estimate the same return to reduce overestimation, rather than evaluating scene-specific and general objectives separately.
This distinction helps explain the observed advantage of SA-MARL under the
tested changing scenarios.

To evaluate the ad-hoc network's resilience, we introduce environmental interference into the tracking scenarios, as shown in Fig. \ref{fig:4d}--\ref{fig:4f}.
Here, the interfered condition refers to environmental disturbances in the tracking scenarios.
Under these interfered conditions, SA-MARL exhibits superior robustness, maintaining stable and ascending learning curves even as the environment becomes more complex.
In contrast, compared algorithms suffer from significant performance oscillations and fail to reach stable convergence, particularly in the large-scale ``12 AUVs tracking 4 targets'' interfered scenario.
The compared methods use their standard policy optimization or actor--critic mechanisms, whereas SA-MARL additionally identifies the current coordination
scene and fuses the scene critic network with the general critic network. The scene critic network emphasizes scene-specific tracking requirements, while the general critic network preserves general coordination and safety objectives.

Following the convergence tests, we evaluate the tracking accuracy of the proposed SA-MARL algorithm using the radial metrics presented in Fig. \ref{fig:5}.
Tracking is considered accurate when the actual AUV--target distance is within 0.5\% of the desired tracking distance. Tracking accuracy is defined as the percentage of evaluated AUV--target pairs satisfying this criterion.
This analysis provides a multi-dimensional comparison of how different algorithms maintain precise target monitoring across varying cluster scales and environmental conditions.
The quantitative results, as illustrated in the circular bar distribution in Fig. \ref{fig:5a}, reveal that the SA-MARL-driven approach consistently outperforms compared approaches across all tested metrics.
In the ``4 AUVs tracking 2 targets'' scenario, the proposed scheme achieves a peak accuracy of 92\%, significantly exceeding the performance of MASAC (74\%) and MAAC (32\%), which highlights the efficacy of the scene-adaptive policy in managing low-density coordination.
This superiority remains evident as the mission scale expands; for the ``12 AUVs tracking 4 targets'' configuration, SA-MARL maintains a robust accuracy of 86\%, while the DSBM scheme falls to 64\%. Furthermore, the proposed approach demonstrates remarkable resilience to unpredictable underwater noise disturbances, preserving an accuracy of 84\% under interfered conditions for the multi-AUV ad-hoc network of 12 AUVs. 
These findings demonstrate that the weight-based dynamic fusion within the Decision Network Layer ensures optimal collaborative synergy by identifying multi-AUV collaborative scenes in real-time.
Consequently, the SA-MARL allows the AUV cluster to maintain stable, high-precision tracking even amidst abrupt topological shifts or intense environmental interference within the ad-hoc networks.

To further assess the robustness of decentralized coordination under communication impairments, we evaluate the influence of packet loss and one-way communication delay on tracking performance, as shown in Fig. \ref{fig:5b}.
As illustrated in Fig.~5(b)(i) and Fig.~5(b)(ii), when the packet-loss rate increases from 0\% to 40\%, the tracking accuracy of SA-MARL decreases from 86.4\% to 65.1\% under the nominal condition and from 84.2\% to 64.3\% under the interfered condition, respectively. 
Although a gradual degradation is observed as packet loss increases, SA-MARL still retains relatively high tracking accuracy across the tested range. In comparison, the influence of communication delay is more limited. As shown in Fig.~5(b)(iii) and Fig.~5(b)(iv), increasing the one-way delay from 0.2 to 1.0~s reduces the tracking accuracy from 84.4\% to 79.8\% under the nominal condition and from 84.1\% to 79.5\% under the interfered condition. 
These results indicate that the proposed SA-MARL strategy can sustain cooperative tracking under the considered communication impairments.

Finally, to further validate the practical effectiveness of the SA-MARL-driven approach in complex underwater environments, we utilize the Unity engine to provide Unity-based 3D visualization of the embodied execution process. The simulation results across three distinct mission scales are illustrated in Fig. \ref{fig:6}.

As presented in Fig. \ref{fig:6}, the embodied execution proceeds through three critical stages, starting with the randomized initialization of AUV clusters and their respective targets across diverse underwater terrains.
During this initial phase, the AUVs begin in a dispersed state and utilize local perception modules to detect coordinates within the 3D operational space.
As the mission transitions into the mid-phase, the cluster engages in a coordinated pursuit where colored trajectories represent smooth paths generated by the scene-adaptive policy, demonstrating the cluster's ability to maintain optimal formations while navigating around complex seafloor obstacles.
In the final stable monitoring phase, the AUV cluster achieves a consistent multi-target tracking configuration where individual AUVs synchronize their velocities with the targets.
Ultimately, the consistency between mid-phase trajectory convergence and terminal stability validates the capacity of our proposed scheme to translate high-level policies into robust low-level control.
By leveraging a dual-path critic and scene-adaptive fusion, the SA-MARL effectively resolves gradient conflicts between global mission objectives and local environmental constraints.
This synergy ensures high-precision tracking and seamless coordination, highlighting our proposed scheme's efficacy for large-scale AUV deployment within bandwidth-constrained ad-hoc networks.

\section{Conclusion}\label{Section:6}
In this paper, we introduced a Scene-Adaptive EI architecture and the SA-MARL algorithm to address the fundamental challenges of multi-target tracking within underwater ad-hoc networks. 
By re-envisioning AUVs as "embodied entities", the proposed framework successfully integrates perception, decision-making, and physical execution into a unified cognitive loop, overcoming the rigidities inherent in traditional data-centric architectures.

The core of our contribution, the SA-MARL algorithm, employs a specialized dual-path critic mechanism and weight-based dynamic fusion.
This design effectively decouples global mission safety from scene-specific coordination, resolving the gradient conflicts typically found in fluid topologies. 
Furthermore, we materialized this framework through a beacon-based communication and control model, bridging the gap between high-level policy inference and decentralized physical actuation in bandwidth-constrained environments.

Extensive evaluations and Unity-based 3D visualizations demonstrate that our scheme consistently outperforms mainstream MARL approaches, achieving significantly accelerated convergence and superior tracking accuracy under both ideal and interfered conditions. 
These results validate the robustness and scalability of the architecture in managing complex marine environments. 
For future work, we plan to extend this EI framework to more diverse underwater missions and investigate its performance in heterogeneous multi-AUV ad-hoc networks with varying physical constraints.
\appendices


\ifCLASSOPTIONcaptionsoff
  \newpage
\fi

\bibliographystyle{IEEEtran}
\bibliography{ref}

\section*{Biographies}

\footnotesize
Kai Tian is currently working toward the bachelor’s degree with the Software College, Northeastern University, Shenyang, China. His research interests include embodied intelligence, deep reinforcement learning, multi-agent reinforcement learning, and machine learning.

\footnotesize
Jialun Wang is currently pursuing a Bachelor’s degree at the Software College, Northeastern University, Shenyang, China. His research interests include reinforcement learning, embodied intelligence, and machine learning.

\footnotesize
Chuan Lin [S'17, M'20] is currently an associate professor with the Software College, Northeastern University, Shenyang, China.
	He received the B.S. degree in Computer Science and Technology from Liaoning University, Shenyang, China in 2011, the M.S. degree in Computer Science and Technology from Northeastern University, Shenyang, China in 2013, and the Ph.D. degree in computer architecture in 2018.
	From Nov. 2018 to  Nov. 2020, he is a Postdoctoral Researcher with the School of Software, Dalian University of Technology, Dalian, China.
	His research interests include UWSNs, industrial IoT, software-defined networking.

\footnotesize
Guangjie Han [S’03-M’05-SM’18-F’22] is currently a Professor with the Department of Internet of Things Engineering, Hohai University, Changzhou, China. He received his Ph.D. degree from Northeastern University, Shenyang, China, in 2004. In February 2008, he finished his work as a Postdoctoral Researcher with the Department of Computer Science, Chonnam National University, Gwangju, Korea. From October 2010 to October 2011, he was a Visiting Research Scholar with Osaka University, Suita, Japan. From January 2017 to February 2017, he was a Visiting Professor with City University of Hong Kong, China. From July 2017 to July 2020, he was a Distinguished Professor with Dalian University of Technology, China. His current research interests include Internet of Things, Industrial Internet, Machine Learning and Artificial Intelligence, Mobile Computing, Security and Privacy. Dr. Han has over 500 peer-reviewed journal and conference papers, in addition to 160 granted and pending patents. Currently, his H-index is 65 and i10-index is 282 in Google Citation (Google Scholar). The total citation count of his papers raises above 15500+ times. Dr. Han is a Fellow of the UK Institution of Engineering and Technology (FIET). He has served on the Editorial Boards of up to 10 international journals, including the IEEE TII, IEEE TCCN, IEEE TVT, IEEE Systems, etc. He has guest-edited several special issues in IEEE Journals and Magazines, including the IEEE JSAC, IEEE Communications, IEEE Wireless Communications, Computer Networks, etc. Dr. Han has also served as chair of organizing and technical committees in many international conferences. He has been awarded 2020 IEEE Systems Journal Annual Best Paper Award and the 2017-2019 IEEE ACCESS Outstanding Associate Editor Award. He is a Fellow of IEEE.

\footnotesize
Shengchao Zhu received his B.S. degree in Internet of Things Engineering from Hohai University, Changzhou, China, in 2023. He is currently pursuing the Ph.D. degree with the Department of Computer Science and Technology at Hohai University, Nanjing, China. His current research interests include swarm intelligence, swarm ocean, Multi-Agent Reinforcement Learning. 

\footnotesize
Ying Liu received the B.S., M.S., and Ph.D. degrees from Northeastern University, Shenyang, China, in 2003, 2006, and 2012, respectively, all in computer science. She is currently an Associate Professor with the College of Software, Northeastern University. She has published over 30 articles and refereed
conference papers. Her current research interests include Service Computing and Edge Computing.

\footnotesize
Qian Zhu is currently an associate professor with the Software College, Northeastern University, Shenyang, China. She received the B.S. degree in Information and Computing Science from Northeastern University, Shenyang, China in 2006,  the M.S. degree in Operation Science and Control Theory from Northeastern University, Shenyang, China in 2008, and the Ph.D. degree in Communication and Information System from Northeastern University, China in 2018. From Nov. 2019 to Nov. 2024, she was a Postdoctoral Researcher with Control Science and Engineering, Northeastern University, Shenyang, China. Her research interests include artificial intelligence optimization algorithms, Unmanned Aerial Vehicle (UAV) technology and software development for applications.

\end{document}